\pdfoutput=1

\documentclass[11pt]{article}

\usepackage[preprint]{acl}

\usepackage{times}
\usepackage{latexsym}
\usepackage{amssymb}
\usepackage{pifont}
\newcommand{\cmark}{\ding{51}}%
\newcommand{\xmark}{\ding{55}}%
\usepackage{booktabs}%

\usepackage{etaremune}

\usepackage[T1]{fontenc}

\usepackage[utf8]{inputenc}

\usepackage{microtype}

\usepackage{inconsolata}

\usepackage{graphicx}

\usepackage{enumitem}

\usepackage{tabu}
\usepackage{multirow}
\usepackage{booktabs}
\usepackage{arydshln}
\usepackage{xspace}
\usepackage{amsmath}

\makeatletter
\renewcommand{\sectionautorefname}{\S\@gobble}
\renewcommand{\subsectionautorefname}{\S\@gobble}
\renewcommand{\subsubsectionautorefname}{\S\@gobble}
\renewcommand{\appendixautorefname}{Appendix \@gobble}
\makeatother

\usepackage{caption} 
\usepackage{subcaption}

\newcommand{\xref}[1]{\S\ref{#1}}

\definecolor{purp}{HTML}{791f87}

\newcommand{\icon}{\raisebox{-1pt}{\includegraphics[width=1.2em]{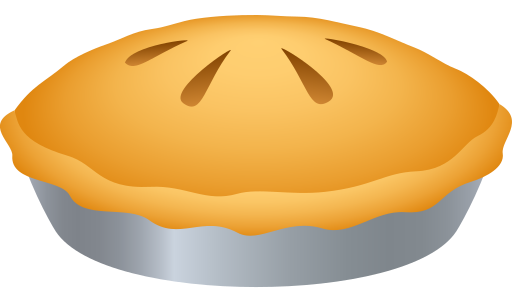}}\xspace}

\newcommand{\llamapie}{\textsc{LlamaPIE}~}

\title{\textsc{LlamaPIE}~\icon: Proactive In-Ear  Conversation  Assistants}

\author{
Tuochao Chen$^\ast$ \quad Nicholas Batchelder$^\ast$ \quad  Alisa Liu$^\ast$ \\
{\bf Noah A.~Smith$^{\ast\dagger}$} \quad  {\bf Shyamnath Gollakota$^\ast$ } \\
{$^\ast$Paul G. Allen School of Computer Science \&\ Engineering, University of Washington} \\ 
{$^\dagger$Allen Institute for Artificial Intelligence}  \\ \url{{tuochao,nicbat,alisaliu,nasmith,gshyam}@cs.washington.edu} 
}

\newcommand{\squishlist}{\begin{itemize}[itemsep=1pt,parsep=2pt,topsep=3pt,partopsep=0pt,leftmargin=0em, itemindent=1em,labelwidth=1em,labelsep=0.5em]}

\newcommand{\squishend}{\end{itemize}}

\begin{document}

\maketitle

\begin{abstract}

We introduce \textsc{LlamaPIE}, the first real-time proactive assistant designed to enhance human conversations through discreet, concise guidance delivered via hearable devices. Unlike traditional language models that require explicit user invocation, this assistant operates in the background, anticipating user needs without interrupting conversations. We address several challenges, including determining when to respond, crafting concise responses that enhance conversations, leveraging knowledge of the user for context-aware assistance, and real-time, on-device processing. To achieve this, we construct a semi-synthetic dialogue dataset and propose a two-model pipeline: a small model that decides when to respond and a larger model that generates the response. We evaluate our approach on real-world datasets, demonstrating its effectiveness in providing helpful, unobtrusive assistance. User studies with our assistant, implemented on Apple Silicon M2 hardware, show a strong preference for the proactive assistant over both a baseline with no assistance and a reactive AI assistant, highlighting the potential of \llamapie{} to enhance live conversations.
\noindent {Code and dataset are available at {\textcolor{blue}{\url{https://github.com/chentuochao/LlamaPIE}}}}.

\end{abstract}

\section{Introduction}

User interaction with language models (LMs) has primarily followed a turn-based dialogue format, where users actively request responses from LM assistants. This approach requires users to shift their attention and carefully phrase prompts to obtain useful answers. Thus, usability is restricted to scenarios where users can pause their activities to engage with the conversational assistant.

\begin{figure}[t!]
    \centering
    \includegraphics[width=\columnwidth]{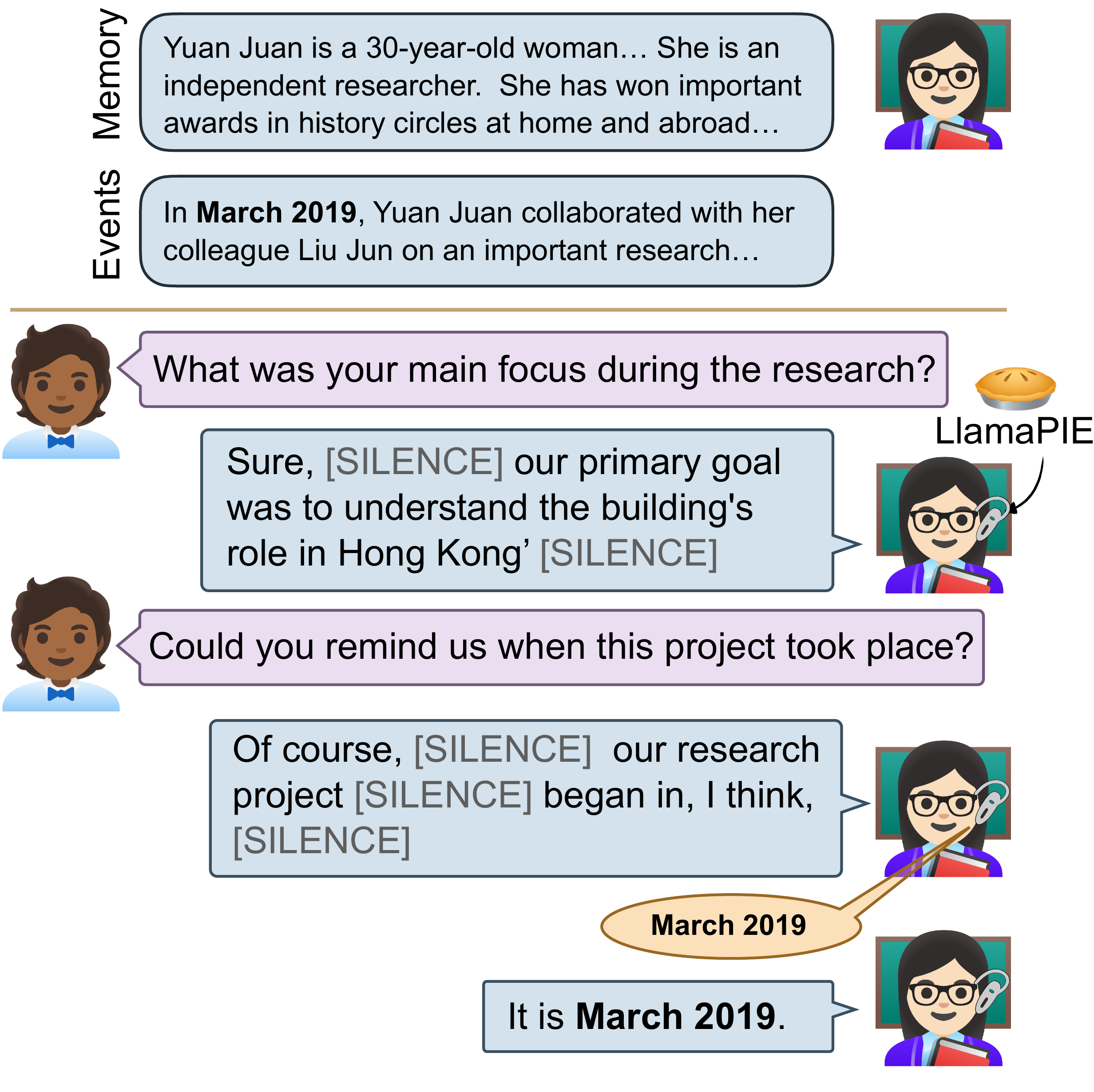}
    \caption{\llamapie is a \textbf{P}roactive, \textbf{I}n-\textbf{E}ar assistant that augments human-to-human communication by providing discreet guidance via hearable devices. Its responses are short, provided only when helpful, and leverage the wearer's memory of past events. {In the figure, LlamaPIE assists Yuan by whispering 1--3 words to her during the conversation only when needed by anticipating user needs, remaining silent most of the time.} }
    \label{fig:figure1}
    \vskip -0.14in
\end{figure}

Here, we explore a compelling alternative: What if an AI system could {\it proactively} assist users without explicit invocation?  We introduce \textbf{\textsc{LlamaPIE}}: proactive  assistants designed to augment human-to-human communication through discreet, context-aware guidance delivered via hearable devices.\footnote{The concise assistance can be delivered audibly through earbuds, headphones, or bone-conduction headsets, or visually via augmented reality wearables like smart glasses.} These real-time assistants  operate as mostly silent co-pilots, providing occasional and unobtrusive feedback to the wearer by anticipating user needs.
Unlike traditional dialogue assistants, they operate in the background and allow users to remain focused on their human interactions.

{
Such proactive companions may enhance human interactions in a variety of contexts, including negotiations~\cite{10.1145/3626772.3657843}, interviews~\cite{mitinterview}, customer support~\cite{10.1145/3626772.3657843}, and cross-cultural communication~\cite{10.1145/3196709.3196734}. They may support neurodiverse individuals, such as those with autism or social anxiety, by assisting in the interpretation of social cues~\cite{10.1145/3654777.3676430}, or serve as personal memory assistants for older adults experiencing progressive memory loss, providing contextually relevant information during conversations~\cite{vimes}. }

Creating proactive assistants presents several challenges. Modern language models are typically designed to provide detailed responses, but in our case, assistance should be limited to just 1--3 words and offered only when it enhances the conversation without disrupting its flow. The assistant must also be context-aware, leveraging the wearer’s memory of prior activities  and interactions. Achieving this enhancement to live conversations requires  real-time streaming processing, while privacy considerations necessitate processing conversational data on edge devices rather than sending it to the cloud.  Finally, there does not exist large-scale annotated data of in situ human conversations with real-time assistance for model training. 

To collect data emulating this kind of assistance, we first construct a semi-synthetic dialogue dataset that incorporates in-ear assistance. 
Each example consists of a user profile, memory, and a dialogue between the user and other speakers with our assistant  aiding the user.
Special silence markers indicate timing information for speaker utterances.

To meet on-device real-time constraints, we design \llamapie as a two-model pipeline that separates the decision-making process into two stages. First, a small classifier determines ``when to respond'' in a streaming manner. Then, when a response is needed, a larger ``what to say'' model generates a concise reply. While both models continuously process input, the large model generates tokens only when triggered by the small model, reducing computational overhead. We tune \textsc{Llama3.2-1b} and \textsc{Llama3.1-8b} on our dataset to create the small and large models, respectively.

To demonstrate the generalizability of our models trained on synthetic data, we evaluate \llamapie on audio recordings  from the MIT interview dataset~\cite{mitinterview}. In automatic evaluation with LLM-as-a-judge, we show that assistance provided is generally helpful and unobtrusive. To validate the LLM-as-a-judge scores, we also perform  annotations with 21 human scorers that show  strong correlation between the human and LM scores.


Finally, we bring \llamapie ``to life'' by integrating it into a speech-to-speech setup that the user wears on a wireless bone-conduction headset. The end-to-end system is implemented on Apple Silicon M2 hardware, supported by commodity mobile devices. All language and speech models operate in a streaming manner, with intermediate states continuously cached to maximize computational reuse and achieve real-time inference. In our study, 15 participants read a passage, then were tested on the reading in an interview setting. With the proactive assistant, user test accuracy rose from 37\% to 87\%. Participants also found \llamapie much less disruptive to the conversation than a reactive system, where users interacted with \texttt{ChatGPT} through both its voice and text interfaces.


Our work demonstrates a vision for proactive LM assistance that centers human-to-human conversation, and highlights the potential for such interactions to serve users in ways that complement the currently widespread human-AI ``chat'' interface.

\section{\llamapie}\label{sec:llamapie}
We first formulate the problem of proactive, in-ear assistance for human conversations (\autoref{subsec:problem_formulation}). Then, we describe the dataset construction (\autoref{subsec:dataset_generation}) and finally our two-model pipeline (\autoref{subsec:proactive_assistant_modeling}).

\subsection{Problem Formulation}\label{subsec:problem_formulation}
The assistant's primary role is to help a single user during a conversation involving multiple human speakers. It has two key functions: proactively determining when to provide assistance and delivering concise, unobtrusive messages. To minimize disruption, responses should be brief (1--3 words) so they do not noticeably interfere with the ongoing discussion. Effective assistance requires the assistant to anticipate the user’s needs, offering help only when necessary while remaining silent most of the time. Additionally, an ideal in-ear assistant should leverage user-specific information, such as past events, to enhance its support.

In this work, we provide the assistant with a natural language ``memory'' (as part of the prompt) containing biographical details about the user and two key events relevant to the current conversation. This context should help the assistant deliver more relevant and timely assistance.

The input consists of an audio stream capturing conversation between two or more human speakers. To process this, streaming automatic speech recognition (ASR) and diarization models can be used to transcribe the audio into text. These models also provide annotations, including speaker identification, timing information for each turn, periods of silence and  speech overlaps. Given our real-time constraints, these models must operate in a streaming manner,  rather than waiting for the entire input.

Similarly, while our in-ear assistant produces speech output, we use text-to-speech (TTS) models for synthesis (details of TTS and ASR models used are in~\xref{sec:inference}). Thus, the core of our system operates with text as both input and output. 


\subsection{Synthetic Dataset Generation}\label{subsec:dataset_generation}

We construct examples by first creating the user memory (\xref{subsec:memory_generation}) and then the dialogue (\xref{subsec:dialogue_generation}).
To improve the diversity and realism of the data, we draw from real conversational contexts and user profile datasets when synthetically generating data.
The assistant should serve two main functions: {\it providing reminders}, e.g., helping users recall secondary details of events like names and places, and {\it social guidance}, e.g., helping the user continue the natural flow of the conversation.
We use \texttt{claude-3-5-sonnet-20240620} with Anthropic API to generate data (see~\xref{sec:dataprompt} for dataset generation prompts). 

\subsubsection{Principles for Proactive In-Ear Assistant}
The core priority for the in-ear assistant is to enhance the user's experience. 
Assistance should be provided in a way that aligns with the user’s immediate focus and  needs. \citet{10.5555/2074094.2074124}  envision an ideal assistant as an intuitive and polite butler—offering useful suggestions when appropriate, delivering real value, and ensuring minimal disruption. To this end, for our in-ear assistant, we draw on the nine principles for proactive behavior outlined by~\citet{intentions}:
\squishlist
    \item \textbf{Valuable}: advances the user's interests and tasks, in the user's opinion
    \item \textbf{Pertinent}: attentive to the current situation
    \item \textbf{Competent}: within the scope of the agent's abilities and knowledge
    \item \textbf{Unobtrusive}: not interfering with the user's own activities or attention, without warrant
    \item \textbf{Transparent}: understandable to the user
    \item \textbf{Controllable}: exposed to the scrutiny and according to the mandate of the user
    \item \textbf{Deferent}: gracefully unimposing
    \item \textbf{Anticipatory}: aware of current and future needs and opportunities
    \item \textbf{Safe}: minimizes negative consequences
\squishend

We use these principles in synthetic dialogue generation in~\xref{subsec:dialogue_generation} and for evaluation  in~\xref{subsec:response_eval}.


\begin{table}[t!]
\caption{Statistics for our semi-synthetic dataset.}
\vskip -0.1in
\centering
{\footnotesize
\begin{tabular}{lrrr}
\toprule
 & Synthetic & SODA & PerLTQA\\
\midrule
\# dialogues & 3128  & 2758 & 3006 \\
\midrule
\textbf{Per-turn statistics} &&&\\
Assistant Length (s) & 0.7 (0.2)   & 0.6 (0.2) & 0.7 (0.2) \\
Speaker Length (s) & 7.5 (3.6)   & 6.7 (3.3) & 8.0 (3.9) \\
Turn interval (s) & 0.5 (0.3)   & 0.5 (0.3) & 0.5 (0.3) \\
Assistant (words) & 2.1 (0.6)   & 2.0 (0.5) & 2.0 (0.6)\\
Speaker (words) & 22 (10)   & 21 (9.5) & 23 (9.9)\\
\midrule
\#non-user speakers & 1.2 (0.7) & 1.3 (0.7) & 1.2 (0.6) \\
Assistant Turns & 4.0 (1.5)   & 3.7 (1.4) & 3.9 (1.5)\\
Speaker Turns & 23 (4.1)   & 22 (4.0) & 22 (4.0) \\
Memory  (words) &79 (7.8)   & 77 (7.9) & 82 (24)\\
Events (words) & 40 (7.0)   & 39 (7.3) & 182 (64)\\
\bottomrule
\end{tabular}
}
\label{tab:syntheticstats}
\vskip -0.15in
\end{table}

\subsubsection{Memory Generation or Selection}\label{subsec:memory_generation}

Each memory instance consists of a structured user profile and two recent events the user has participated in.
The memory comes from one of three distinct sources: synthetic memory, created from predefined keywords; SODA-based memory, derived from the contextual setup preceding a dialogue in the SODA dataset~\cite{kim-etal-2023-soda}; and PerLTQA-based memory, which consists of direct memory samples from the PerLTQA dataset~\cite{perltqa}. Statistics are shown in \autoref{tab:syntheticstats}. SODA is a large-scale social dialogue dataset covering a wide range of social interactions, while PerLTQA is a personal long-term memory dataset designed for question-answering tasks.

For synthetic memory, we use Claude to generate a user profile by providing five randomly selected keywords from a predefined list of 100 (see \autoref{sec:keywords}). Given these keywords, the model constructs a user profile and corresponding events. For SODA-based memory, instead of keywords, the model uses the context preceding SODA dialogues to generate the user profile. For PerLTQA-based memory, we select a random profile from the PerLTQA dataset, and two random events attached to that profile.


\subsubsection{Dialogue Generation}\label{subsec:dialogue_generation}
The next step is dialogue generation, grounded in the user memory. We instruct Claude to construct a scenario that the user might encounter given their background. Each scenario is designed to fit a specified scenario type and illustrate a predefined use case, while also exemplifying two randomly selected principles of proactive assistants. To account for scenarios where users ignore  assistance,  some dialogues include an explicit instruction for Claude to generate interactions in which the user occasionally ignores the assistant’s messages (\autoref{sec:dataprompt}).

When generating dialogue based on SODA-derived memory, Claude is seeded with the first three lines of the corresponding SODA conversation for continuity with the original context. 

All generated dialogues are structured with timestamps marking the start and end of each sentence, along with speaker identifiers such as \texttt{User}, \texttt{Speaker \#N}, or \texttt{Assistant}. Turns between speakers  have a random gap chosen between --1 and 1 second, allowing for overlaps and silences between turns. Additionally, dialogues can incorporate hesitation markers, formatted as ``(hesitation $n$ ms)'', enabling more natural pauses in speech patterns. 

\vskip 0.04in\noindent{\bf Formatting  for streaming and timing.} To enable tokenized streaming and preserve timing information in dialogues --- details typically absent in natural language data used for training LMs --- we reformat Claude-generated dialogues to accommodate real-time processing. Gaps between speakers are replaced with silence markers, \texttt{|SILENCE >}, with each token representing 0.5 seconds of silence. Generated hesitation markers are replaced with silence markers as well (the hesitation duration is rounded up to the nearest equivalent number of tokens). 
This ensures that pauses and hesitations are accurately represented in the dataset, even when Claude’s  generation  does not include sufficiently long hesitations. Furthermore, since all gaps and hesitations are converted into the same silence marker, the streaming models do not have explicit marked information about hesitations. Future work could focus on more sophisticated modeling of conversational turn-taking, potentially leading to more realistic synthetic data.

\subsection{Proactive Assistant Modeling}\label{subsec:proactive_assistant_modeling}
Our proactive assistant operates in a streaming manner by (1) processing the output from the speech models, (2) predicting whether the user requires assistance, and (3) generating the appropriate response. All of these operations must run efficiently on a local device under real-time constraints.

\begin{figure}
    \centering
    \includegraphics[width=\columnwidth]{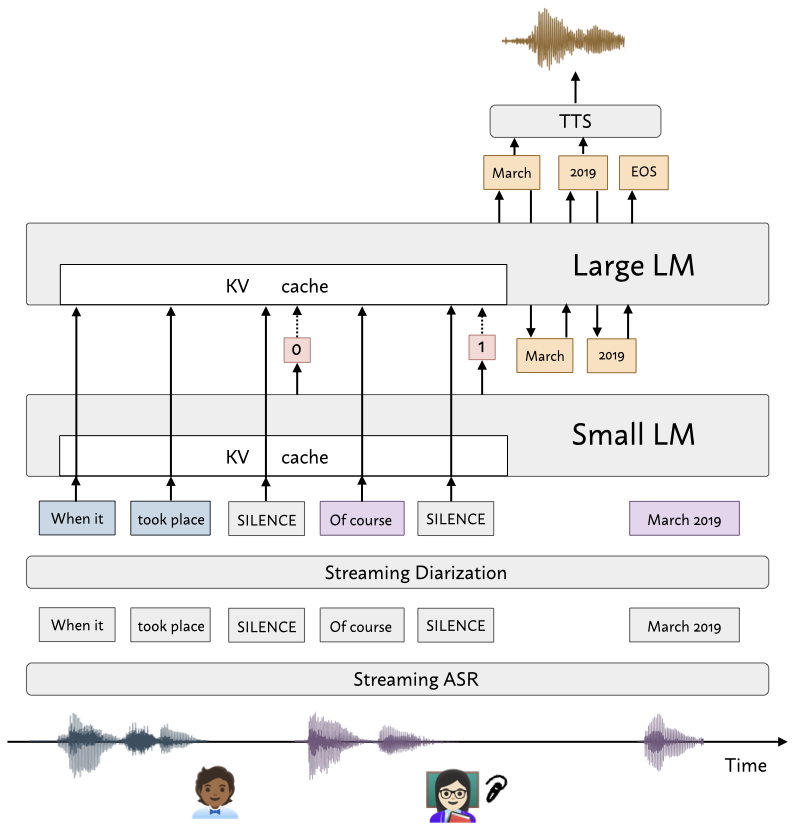}
    \caption{Illustration of our dual-model pipeline.}
    \label{fig:architecture}
    \vskip -0.13in
\end{figure}

\subsubsection{Dual-Model Architecture}
Autoregressive LMs predict the next token conditioned on previously generated tokens. This sequential generation introduces inherent latency, making real-time or low-latency applications challenging. To address this inefficiency, dual-model strategies have been proposed, such as  speculative decoding~\cite{10.5555/3618408.3619203}, for normal text generation tasks. Following a similar intuition, we  design a dual-model strategy for our streaming inference. {Our key computational overhead arises from performing continuous inference on incoming tokens. Thus, we employ a small model to continuously process these tokens in a non-autoregressive manner and determine when the user requires assistance, as shown in Fig.~\ref{fig:architecture}.} When the small model triggers the assistance, the larger model is called to generate the response in an autoregressive manner. In this way, we leverage the high efficiency of the small model to run and predict continuously, and the stronger capabilities of the large model to generate a high-quality response.

To further reduce false positives caused by the small model, the large model is trained to generate an end-of-sequence (EOS) token if it determines that the user does not actually need help, even if the small model initially triggered assistance. When the large model generates the response, we also stream the assistant's response back into the pipeline so that both small and large models are aware of the provided assistance. Empirically, we found that if the small and large model are not aware of assistance history, they will try to generate repeated assistance multiple times at nearby places in the conversation. 
    
\subsubsection{Model Finetuning}
We tune \textsc{Llama3.2-1B-Instruct} for our small model and \textsc{Llama3.1-8B-Instruct} for our large model, both using LoRA~\cite{hu2022lora}. 
For the small model, we add a classification head on top of the last-layer output. We only train the model to make a prediction on silence markers. We tune the large model using 75\% positive examples and 25\% negative examples. Positive examples are randomly selected dialogue locations where the assistance responds, and the model is trained to generate the target response. Negative examples are randomly chosen positions with no assistant assistance, at least two conversation turns away from assisted positions. For negative examples, the model is trained to output the EOS token.

\subsubsection{On-Device Real-Time Inference}\label{sec:inference}

We employ the streaming ASR model from SpeechBrain~\cite{speechbrainV1} to transcribe speech into text in real time, using a chunk size of 960~ms and a context length of 3840~ms. We use the streaming diarization model from Diart~\cite{diart} to detect speaker turns within the conversation. For text-to-speech synthesis, we implement the FastSpeech2 model~\cite{fastspeech} to generate and play back audio. All speech models run on the  CPU. The ASR model processes a 960 ms chunk in 20.4 ms, the Diart model in 6 ms, and the TTS model converts 1–3 words in 37 ms on average.

We implement our dual Llama models using the MLX framework~\cite{mlx2023} on the MPS device of Apple Silicon. The small model is quantized to \texttt{bfloat16}, while the large model is quantized to \texttt{int8} for optimized performance. During streaming inference, both models continuously cache KV states for computational reuse. We use the testbench from MLX framework to measure the runtime and memory of dual Llama models in Apple M2 chip with 16GB memory. The token processed speed of the small model is  38.7 tokens/s and peak memory consumption is  2.49GB. The token generation speech of the large model is  14.2 tokens/s and the peak memory is 8.9GB. Our dual-model architecture achieves at least a 64\% reduction in processing time during continuous inference compared to the implementation of a single large model.

\section{Evaluation}
\label{sec:experiments}
We first discuss evaluation metrics \autoref{subsec:metrics}, followed by results on the system's ability to decide when to respond \autoref{subsec:accuracy_eval} and the quality of its responses \autoref{subsec:response_eval}.


\subsection{Metrics}\label{subsec:metrics}
\subsubsection{Quantitative Metrics}


To first measure whether \llamapie responds when it should, we compute its precision, recall, and accuracy (P/R/A) on our synthetic test set, using the data as ground truth for when assistance should be provided. 
In addition to hard P/R/A, we also report soft P/R/A, which gives model leniency of $\pm 1$ turn to respond, when the ground-truth data contains an assistance response.
This is because responses can often be equally helpful when provided at many different nearby points in a conversation.





\subsubsection{Qualitative Metrics}\label{subsubsec:qualitative_metrics}
Due to the costly nature of large-scale human evaluation, we follow recent work and use LM-as-a-judge \cite{10.5555/3666122.3668142}.
To improve the validity of our LM evaluator, we use several key strategies. First, since we use \texttt{Claude} to generate our data, we used \texttt{GPT-4o} as our evaluator. Second, we apply a score rubric to our evaluation prompting, following suggestions from prior work~\cite{kim2023prometheus}. 

Finally, we also ensure that our evaluator has high correlation with human judgment through a human annotation experiment.

\begin{table}[t!]
\caption{Small model accuracy at predicting when to respond.}
\vskip -0.1in
\centering
{\footnotesize
\begin{tabular}{lcccc}
\toprule
Metric & Synthetic & SODA & PerLTQA\\
\midrule
Hard Precision & 0.757 & 0.728 & 0.759\\
Hard Recall & 0.719 & 0.727 & 0.777 \\
Hard Accuracy & 0.935 & 0.939 & 0.932\\\midrule
Soft Precision & 0.937 & 0.906 & 0.900\\
Soft Recall & 0.889 & 0.921 & 0.903 \\
Soft Accuracy & 0.978 & 0.977 & 0.976 \\
\bottomrule
\end{tabular}
}
\label{tab:smallmodelaccuracy}
\vskip -0.15in
\end{table}

\begin{table*}[t!]
\caption{Evaluation of \llamapie{} on synthetic and real-world datasets (SODA, PerLTQA, MIT). We report the mean (standard deviation) of scores on a scale from 1 to 5, assigned by \texttt{GPT-4o} (see \autoref{subsubsec:qualitative_metrics} for a description of the metrics). Responses are generally rated highly on all principles.}
\vskip -0.1in
\centering
{\footnotesize
\begin{tabular}{lcccc}
\toprule
 & Synthetic & SODA & PerLTQA & MIT\\
\midrule
\textbf{Response Stats} &&&&\\
\midrule
Response Frequency & 14\% & 14\% & 15\% & 5.8\%\\
Word Length & 2.08 (0.54) & 2.06 (0.61) & 2.05 (0.60) & 2.03 (0.49) \\
\midrule
\textbf{Nine Principles Scoring ($\uparrow$)} &&&\\
\midrule
Valuable & 4.32 (1.01) & 4.28 (0.96) & 4.22 (1.10) & 4.34 (0.70)\\
Pertinent & 4.52 (0.94) & 4.55 (0.92) & 4.32 (1.14) & 4.77 (0.62) \\
Competent & 4.67 (0.81) & 4.73 (0.79) & 4.60 (0.97) & 4.92 (0.34)\\
Unobtrusive & 4.77 (0.61) & 4.82 (0.60) & 4.73 (0.67) & 4.80 (0.50)\\
Transparent & 4.73 (0.73) & 4.77 (0.75) & 4.62 (0.95) & 4.92 (0.37)\\
Controllable & 4.74 (0.71) & 4.74 (0.77) & 4.65 (0.96) & 4.88 (0.44)\\
Deferent & 4.77 (0.68) & 4.79 (0.71) & 4.22, (1.14) & 4.83 (0.52)\\
Anticipatory & 4.35 (0.99) & 4.30 (0.94) & 4.22 (1.14) & 4.22 (0.71)\\
Safe & 4.89 (0.42) & 4.85 (0.55) & 4.82 (0.55) & 4.94 (0.31)\\
\midrule
\textbf{Rubric Score ($\uparrow$)} & 4.21 (1.20) & 4.19 (1.12) & 3.94 (1.31) & 3.68 (0.98)\\
\bottomrule
\end{tabular}
}
\label{tab:big_results}
\vskip -0.15in
\end{table*}

{\bf Validation with human judgment.} To assess the quality of assistant responses, we evaluate each individual assistant response within a dialogue, comparing ratings between \texttt{GPT-4o} and human annotators. \texttt{GPT-4o} is instructed to assign a score to each response, while human annotators perform the same task based on a predefined rubric. Both LM and human annotators classify responses into one of five categories: (1) not relevant/not used, (2) relevant but redundant/not needed, (3) relevant but not acted on, (4) relevant but used later, and (5) highly relevant/immediately used. These categories are worded to reduce ambiguity between how the evaluators rate responses. The full description of this rubric is shown in ~\xref{subsubsection:rubric}. To mitigate hallucination and overly optimistic scoring, \texttt{GPT-4o} is also required to explain its reasoning in terms of both relevance and timeliness.


We sample 120 dialogues from our synthetic dataset, with equal representation from the categories, synthetic, SODA, and PerLTQA. Each dialogue contains 1–9 proactive assistant responses. These dialogues are divided into 24 forms, with 21 human annotators scoring at least two forms each such that each dialogue is evaluated by two different annotators.

To measure agreement between human and LLM evaluations, we compute the Pearson correlation coefficient between human annotators for each sample and compare it to the correlation between \texttt{GPT-4o} and a randomly selected human rating. The Pearson coefficient measures the linear correlation between two datasets, making it suitable for our 1–5 scoring system, where we expect a strong correlation.

We find that human-LM correlation is stronger with $r=0.652$ compared to human-human correlation with $r=0.636$.
Human validation is conducted only for the rubric scores, not for all nine principles, due to the time-intensive nature of human annotations. 
Nonetheless, the overall score for each assistant response correlates strongly with LM assessment.

{{\bf Assessing quality of synthetic datasets.} We assessed the quality of our synthetic Claude-generated datasets using the above human validated GPT-based evaluator.  The synthetic datasets achieved  high rubric scores of  4.77, 4.78 and 4.88 for Synthetic, SODA, and PerLTQA, respectively.  Across these datasets, the distribution of rubric scores was 0.68\% with a score of 1, 0.51\% with a score of 2, 6.83\% with a score of 3,  1.45\% with a score of 4 and 90.51\%  with a  score of 5.}



\subsection{Small Model Accuracy Evaluation}\label{subsec:accuracy_eval}
We evaluate the small model's ability to anticipate when it should respond. Table~\ref{tab:smallmodelaccuracy} shows that across all three dialogue test sets, hard precision and recall exceed 70\%, while hard accuracy surpasses 93\%. Additionally, soft precision and recall are around 90\%, with soft accuracy exceeding 97\%. These results indicate that the finetuned small model effectively anticipates when it should respond and that allowing a ±1 turn flexibility in response timing improves  recall and precision.

\subsection{Dual-Model Evaluation}\label{subsec:response_eval}
We first evaluate our dual-model pipeline on the test set of generated synthetic dialogues. Each type of synthetic dialogue dataset contains 100 samples. As shown in \autoref{tab:big_results}, the average response word length across all three datasets is approximately two words, and the response frequency is around 15\%. The rubric scores for synthetic, SODA, and PerLTQA dialogues are 4.21, 4.19, and 3.94, respectively. The table also presents the scores for each evaluation principle across these datasets.

We also assess our pipeline on the MIT Interview dataset~\cite{mitinterview}, which consists of real-world mock interview recordings of MIT students seeking internships as they interact with professional career counselors. We randomly select 100 conversation samples, transcribe the audio using our ASR and diarization model, and feed the text into our dual-model inference pipeline in a streaming manner. As shown in Table~\ref{tab:big_results}, the response word length remains around two words, while the rubric score is 3.68.

\begin{table*}[t!]
\vskip -0.07in
\caption{Ablation Study on MIT interview dataset. {We evaluate on 100 conversation samples from the MIT dataset, consisting of a total of 1,324 turns. `Triggered by small' represents the number of instances where the small model attempts to trigger assistance. `Responded by large' represents the number of instances where the larger model provides actual assistance. Note that the `Triggered by small' metric for the assistance-aware model is influenced by the large model, as the generated assistance is streamed back to the small model. The row with * corresponds to the results of the MIT interview dataset in Table.\ref{tab:big_results}}}
\vskip -0.1in
\centering
{\footnotesize
\begin{tabular}{llcccc}
    \toprule
    Small model  & Large model & Large model & Triggered  & Responded  & Rubric score \\
    assistance-aware & config & negative proportion & by small  & by large &  \\
    \hline
    \xmark & 8b finetuned & 0\% & 530 & 530 & 3.38 (0.83)\\
    \xmark & 8b finetuned & 25\% & 530 & 105 & 3.38 (0.93) \\
    \cmark & 8b finetuned & 0\% & 180 & 180 & 3.59 (0.93)\\
    \cmark & 8b finetuned & 25\% & 222 & 101 & \textbf{3.68 (0.98)}*\\
    \cmark & 8b finetuned & 50\% & 283 & 28 & 3.43 (0.88)\\
    \hline
    \cmark & 1b finetuned & 25\% & 226 & 120 & 3.39 (0.84)\\
    \cmark & 8b prompting & -- & 185 & 185 & 3.15 (0.91)\\
    \bottomrule
\end{tabular}

}
\label{tab:ablation}
\vskip -0.15in
\end{table*}

Interestingly, the response frequency is significantly lower at approximately 5.8\%. Unlike our synthetic datasets, which are designed with proactive assistance in mind, the MIT interviews involve natural conversations where users do not expect an in-ear assistant. This demonstrates that the proactive assistant often remains silent when assistance is not anticipated, adapting appropriately to real-world scenarios.  

\subsection{Ablation Study}
We conduct ablation studies across different small and large model configurations on the MIT interview dataset. For the small model, we have two configurations: one finetuned to be  aware of prior assistance provided  by the assistant and the other is not aware of prior assistance. For the large model, we compare five different configurations: (1) \texttt{Llama3.1-8B} finetuned with no negative samples, (2) \texttt{Llama3.1-8B} finetuned with 25\% negative samples, (3) \texttt{Llama3.1-8B} finetuned with 50\% negative samples, (4) \texttt{Llama3.1-1B} finetuned with 25\% negative samples, and (5) \texttt{Llama3.1-8B} with prompting and no  fine-tuning. As shown in \autoref{tab:ablation}, if the small model is not aware of prior assistance, it will  trigger assistance 2--3$\times$ more often. Moreover, being history-aware improves the overall response quality (rubric score). When we compare different large model configurations, we found that simply prompting the \texttt{8B} model or finetuning the \texttt{1B} model are both worse than finetuning \texttt{8B}, showing that both the larger model size and finetuning are useful ingredients. Finally, the \texttt{8B} model finetuned with 25\% negative samples gave the highest rubric score, and we use this as our final model.

{
\subsection{Large Model with Manual Triggering}
One benefit of our dual-model pipeline is that it can be  adapted for manual triggering by replacing the small model with user input. We hypothesize that the large model would continue to generate relevant and appropriate assistance when triggered manually. To simulate manual triggering and evaluate generalization, we removed the small model and programmatically triggered the large model at the ground-truth positions in the synthetic dataset. We then ran the GPT evaluator on these responses. The rubric scores were 4.31, 4.38, and 4.10 for Synthetic, SODA, and PerLTQA, respectively—an average improvement of 0.15 compared to when the small model is used in \autoref{tab:big_results}.}

\section{User Study: \llamapie in Real-Time Human Conversations}

To evaluate \textsc{LlamaPIE}'s potential in assisting real users during live conversations, we use our  pipeline on a MAC M2 platform with 16 GB of memory. Speech assistance is delivered via a Shokz OpenMove bone-conduction headset, which do not obstruct the user's hearing of the conversation.
We recruit 15 human participants (ages 20–40) to interact with our real-time on-device prototype and assess their experience. To simulate conversations where users may require assistance, we design mock interview and trivia scenarios. Each scenario consists of eight topics, with five questions per topic—two easy and three difficult.

{ Each participant experiences three different conditions: {\bf Control}, where no assistance is provided; {\bf Proactive (ours)}, where users receive assistance from \llamapie through the  headset; {\bf Reactive (baseline)}, where users can access \texttt{GPT-4o} via a web UI during the conversation.
In addition, five participants also experienced the following condition: {\bf Reactive (short audio)}, where users  access \texttt{GPT-4o} via its voice mode during the conversation. \texttt{GPT-4o} was prompted to respond concisely (1--3 words) and users received assistance through the headset.}

For each participant, we randomly assign a different topic to each of the three conditions. All participants begin with the control condition, while the order of the baseline LM and proactive assistance conditions is randomized. Participants are given 3--5 minutes to read and memorize the background information for their assigned topic. The background information falls into two categories: (1) Wiki-style descriptions of information-dense topics such as quantum mechanics and DNA computing, and (2) Detailed profiles of fictional individuals, including their careers, families, hobbies, and personal interests. Further details are  in~\autoref{sec:detailsofuserstudy}. 

Following the reading period, participants engage in a casual conversation, during which they are asked one easy question and two difficult questions. A transcript of the spoken dialogue, along with timestamps, is recorded for analysis.

\subsection{Results}
We collect a total of 50 dialogues. The average number of turns per dialogue is 11.2$_{\pm 1.97\,\text{(std)}}$. 
The average number of words per assistant response is 1.83, and the average response frequency is 25\%. 

\vskip 0.01in\noindent{\bf Accuracy.} In  control, participants achieved 37.0\% accuracy on the prepared questions. Accuracy improves to 88.9\% with Reactive \texttt{GPT-4o} assistance and to 86.7\% with our Proactive assistance. 

{
\vskip 0.01in\noindent{\bf Reaction time.} We measure  the interval between when a speaker finishes asking a question and when the user actually starts to answer the question. This includes conversation gaps,  thinking time and hesitation/filling words at the beginning of response. Participants have the shortest average reaction time of 3.29$_{\pm 2.28}$ seconds in the control condition. With reactive \texttt{GPT-4o} assistance, the reaction time increases to 13.38$_{\pm 10.23}$ seconds {and 11.67$_{\pm 9.27}$ seconds for {\bf Reactive (baseline)} and {\bf Reactive (short audio)}, respectively.}  Proactive assistance maintains a lower reaction time of 4.89$_{\pm 3.55}$ seconds.} 

\begin{figure}[t!]
    \centering
    \vskip -0.16in
        \begin{subfigure}[t]{0.52\columnwidth}
    \includegraphics[width=\linewidth]{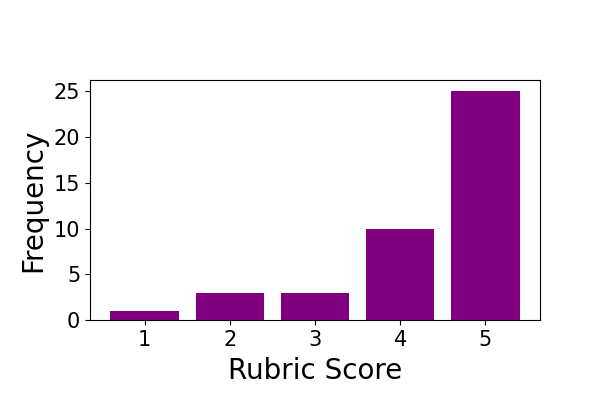}
    \vskip -0.1in
    \caption{}
        \end{subfigure}%
        \begin{subfigure}[t]{0.52\columnwidth}
        \includegraphics[width=\linewidth]{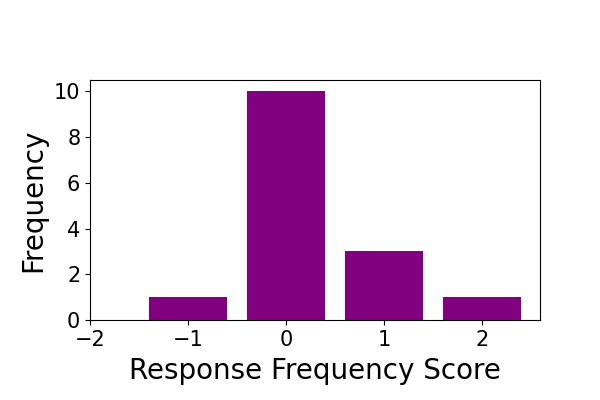}
        \vskip -0.1in
    \caption{}
    \end{subfigure}
    \vskip -0.15in
    \caption{(a) shows the histogram of rubric score across all participants and (b) shows the histrogram of response frequency scores rated by the participants. }
    \label{fig:histogram}
    \vskip -0.18in
\end{figure}

\paragraph{Perceived assistance quality.}
On the rubric score from~\xref{subsubsection:rubric}, participants give an average rating of 4.31$_{\pm 1.03}$. 
To gain deeper insights into response frequency, for Fig.~\ref{fig:histogram}(b), we ask participants to rate it using an opinion scale  from -2 to +2:
\squishlist
  \item  --2: $>1$ assistance  was  unnecessary
  \item  --1: One assistance instance was unnecessary
   \item  0: The amount of assistance was appropriate
  \item   +1: One more assistance instance was needed
  \item   +2: $>1$ additional assistance  was needed
\squishend


The Mean Opinion Score (MOS) is 0.27$_{\pm 0.68}$, indicating  users generally found the response frequency to be appropriate, with slight variation in individual preferences.

Finally, we ask participants to compare how the reactive and proactive assistants affect conversation flow. 
They rate the level of disruption on a scale from 1 to 5, where a score of 1 was ``Strongly disagree that the system was disruptive'' and  5 was ``Strongly agree that the system was disruptive''.  The Mean Opinion Score (MOS) for the reactive baseline system is 4.73$_{\pm 0.57}$, indicating high perceived disruption. In contrast, the MOS for our proactive assistance is 2.4 (standard deviation: 1.2), demonstrating lower  impact on conversation flow. {Participants noted that  {\bf Reactive (short audio)}   exacerbated disruption, as their queries became audible to others, further interfering with  conversations.}

Overall, we find that human performance with \llamapie is on par  to using a state-of-the-art LM assistant, while overwhelmingly preserving the natural flow of the conversation as measured by both response latency and perceived disruption.

\section{Related Work}

\noindent{\bf Language model post-training.} 
The release of ChatGPT popularized a particular setting for user-AI interaction, where users actively seek assistance from AI assistants in dialogue form.
The process of adapting pretrained-only language models to these ``chatbots'' is known as post-training, and has been the focus of much NLP research in the last two years, with key advances in algorithms \cite{schulman2017proximalpolicyoptimizationalgorithms, rafailov2023direct, meng2024simpo}, datasets \cite{wang-etal-2023-self-instruct}, and understanding of the effect \cite{gudibande2023false, lin2024the, hewitt2024instructionfollowinginstructiontuning}. Prior work also explores techniques to improve the response quality of these chatbots by  clarifying user intent with multi-turn interactions~\cite{10.1145/3471158.3472232,10.1145/3404835.3462839, 10.24963/ijcai.2023/738,qian-etal-2024-tell} or initiating peer support chats based on user history~\cite{10.1145/3654777.3676430}.

\vskip 0.02in\noindent{\bf Spoken dialogue models.}
Recent work on spoken dialogue research covers  dialogue state tracking~\cite{monet} and turn-taking prediction~\cite{turn_taking}. Speech-to-speech  models~\cite{nguyen-etal-2023-generative,moshi,syncllm} enable interruptions and real-time adjustments to  sound more human-like. However, these dialogue  models  require explicit user engagement and initiation and  are neither designed to operate in the background  nor to proactively  enhance  human-human conversations.

\vskip 0.02in\noindent{\bf Proactive task planning.} Prior work explores proactive  assistants for task refinement~\cite{zhang-etal-2024-ask}, text-to-SQL support~\cite{wu-etal-2024-need}, and predicting tasks (e.g., sending an email) based on  history like keyboard, mouse,  and web interactions~\cite{lu2024proactiveagentshiftingllm}.  While related, these  are not speech assistants and do not enhance conversations.

\vskip 0.02in\noindent{\bf Life-logging and memory augmentation.} Prior work proposed   devices  that capture audio and video signals for memory augmentation and recall~\cite{prosthesis,10.5555/1037465,vimes}. These systems primarily function as retrieval-based tools, recording audio and enabling keyword-based search~\cite{vemuri} and browsing via smartphone apps~\cite{6152455}. Further, \citet{10.1145/3196709.3196734} identify keywords in conversations and display relevant search results on a visual interface, while~\citet{10.1145/3544548.3581566} create context-aware pictures. 

Recent work~\cite{memoro} uses LMs and prompting for  memory augmentation by having the user explicitly initiate assistance either via a question or a push button. This requires redirecting the user's attention to pause and engage/initiate the AI. In contrast, our work is the first to design  and evaluate a fully proactive  in-ear assistant that  does not require   explicit user invocation. 



\vskip 0.02in\noindent{\bf Human-centric design.} Prior  research  explores user expectations of proactive systems. \citet{miksik2020buildingproactivevoiceassistants} suggest that next-generation devices should deliver timely, relevant information proactively. Wizard-of-Oz studies~\cite{10.1145/3543829.3543834} reveal that while users value proactivity, they have concerns about agency loss and intrusiveness. \citet{intentions} and \citet{linguistic} provide linguistic-driven design guidelines for a proactive assistant design emphasizing unobtrusiveness, safety, and relevance. These guidelines informed the design of our  proactive  assistant.

\section{Conclusion}

We introduce the first proactive in-ear conversation assistant, designed to provide discreet, concise guidance. Our proof-of-concept, real-time, on-device implementation, which centers human-to-human conversation, demonstrates the potential for such interactions to serve people in ways that are complementary to the currently widespread human-AI ``chat'' interface. Our experimental findings show the promise of such tools and establish a starting point for future research.

\section{Limitations and Risks}

\textbf{Limitations.} Currently, we provide memory only in the form of text for the user profile and  prior conversations and events. Further research is needed to develop a system that automatically manages memory based on speech inputs (and potentially other sources). The models in our real-world prototype are trained solely on synthetic datasets; performance could likely be improved with real-world datasets, which are currently lacking. Our real-time prototype could eventually help to generate such datasets with human annotations. We have integrated open-source streaming ASR and TTS models into our dual-model pipeline to create a speech-to-speech setup. Instead of using cascaded systems, future work could integrate multimodal speech models that directly process speech tokens to reduce latency. Additionally, incorporating personalization and human feedback could further enhance the customization of the in-ear assistant.

\noindent\textbf{Ethical considerations.} 
Real-time in-ear proactive assistants have the potential to enhance human conversations by providing discreet, context-aware support. In the future, they could assist in complex interactions such as negotiations, interviews, customer service, and cross-cultural communication. They may also support neurodiverse individuals, such as those with autism or social anxiety, by helping interpret social cues, as well as aid individuals with cognitive challenges, caregivers, and those with high workloads or sensory impairments.  

It is also crucial to discuss potential risks: 
For instance, there are risks of misuse, such as cheating in exams,  or other scenarios where an in-ear assistant would not be ethical. Since our in-ear assistant requires visible wearables like earbuds or headsets, their presence can serve as a potential indicator to mitigate abuse. We encourage future research on the impact of AI technology for aiding human conversation.

In our work, we assume consent from all parties in the conversation. In the US, 38 states  have a ``one-party'' consent requirement for recording conversations~\cite{law}, while the others require consent from all participants in the conversation. Potential solutions include default opt-out and opt-in options based on speech characteristics and consent and the ability to delete data upon request.

\bibliography{references}

\appendix






\section{Generated Speech Duration}
\begin{figure}[t!]
    \centering
    \includegraphics[width=0.85\columnwidth]{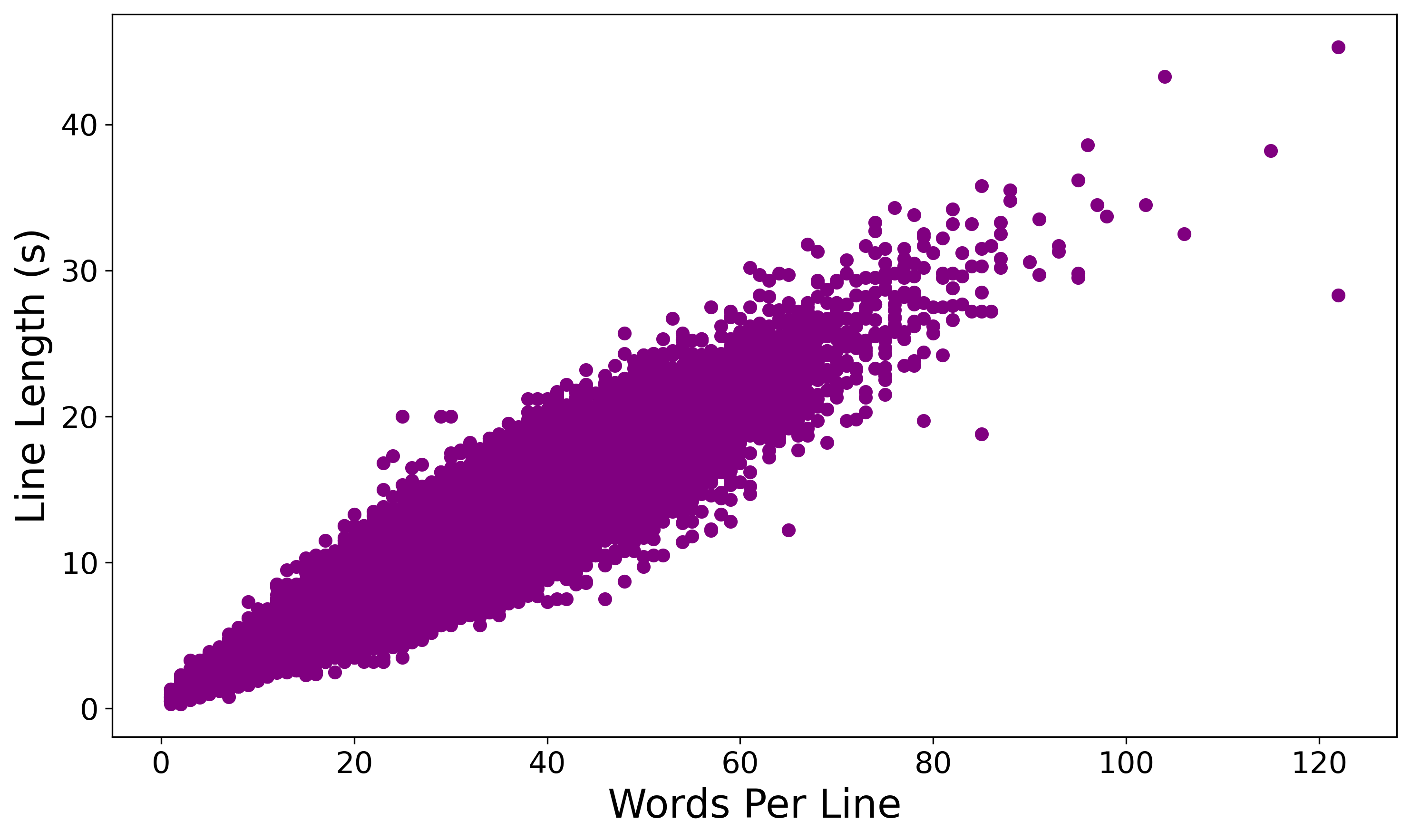}
    \vskip -0.15in
    \caption{Scatterplot representing line length in words vs. line length in seconds. This demonstrates that for sentences generated by Claude, sentence length in words is proportional to sentence length in seconds.}
    \label{fig:linelength}
    \vskip -0.18in
\end{figure}

Understanding the relationship between sentence length in words and sentence duration in seconds helps ensure realistic dialogue generation. As shown in \autoref{fig:linelength}, there is a strong correlation between these two measures, suggesting that Claude has some  implicit sense of how long it takes to say a given sentence. This relationship is important for generating natural conversations, as it indicates that the model may also have some understanding of timing between speakers and how dialogue unfolds over time.

\section{Additional Finetuning Details}

The small model is fine-tuned with an initial learning rate of 2e-5 for 10 epochs with batch size 8 on one A100. We finetune the large model with an initial learning rate of 2e-5 with batch size 16 on 4 L40s for 3 epochs. 
To account for ASR errors from the speech model, we apply data augmentation for more robust finetuning, including: (1) random word dropout with a 2\% drop rate, (2) random word flipping with a 3\% flipping rate, and (3) phonetically similar word replacement with a 1\% replacement rate. We select the checkpoints with best val loss to evaluate on the test set.

\section{User study participants and annotators}
The study was approved by the UW IRB. All participants provided consent and were recruited from our institution and nearby areas. They were offered a \$15 compensation. Participants ranged from 18 to 57 years old, with 35\% identifying as female and the rest as male.

\section{Additional Data Generation Details}

\subsection{Example Data Format}\label{sec:dataformat}

\subsubsection{Memory and Profile of User}
{
{\bf Memory:} Xiao Ming is a 28-year-old male. He is an environmental engineer with extensive work experience. He is from China and has features of black hair, brown eyes and medium build. He loves the outdoors, photography and traveling. His achievement is receiving an award for outstanding research achievements in the field of environmental protection. His ethnic background is Asian. Educational background is in environmental engineering. The employer is an environmental protection technology company. Awards and achievements include being influenced by environmental expert Wang Ming and dedicated to contributing to environmental protection.

{\bf Event 1:} Yang Juan is Xiao Ming's mother. She has a strong interest in astronomy. While Xiao Ming was studying environmental engineering, he and his mother often discussed astronomy-related topics. One day, Yang Juan mentioned Copernicus' 'On the Movement of Celestial Bodies' to Xiao Ming. She explained in detail Copernicus's heliocentric view and the challenge to the geocentric view. Yang Juan was deeply inspired by Copernicus's works and expressed her appreciation for his courage and scientific spirit. She also hoped that Xiao Ming could also make breakthroughs in the field of environmental engineering. This discussion aroused Xiao Ming's keen interest in the history and scientific development of astronomy, further deepening his love for scientific research. Xiao Ming decided to draw on Copernicus' concepts and methods in his research, commit himself to contributing to environmental protection, and become an excellent environmental engineer in the future.

{\bf Event 2:} Together with his neighbor Liu Lin, Xiao Ming planned a trip to the Cocos Islands Marine Reserve. As an environmental engineer, Xiao Ming is particularly interested in marine ecological protection and is passionate about exploring and understanding protected areas around the world. He learned that the marine protected area of Cocos Islands was an important oceanographic case in the geographical field, so he invited his neighbor Liu Lin to go with him. Liu Lin also loves the natural environment and travel. She has heard about the beautiful scenery of the Cocos Islands for a long time and is looking forward to this trip. Together, they plan to set off on May 12, 2023, to experience first-hand the spectacular scenery of the Cocos Islands Marine Reserve through diving and snorkeling activities. Xiao Ming and Liu Lin will interact with various coral reefs and marine life, and gain an in-depth understanding of the importance of marine ecological protection under the guidance of professional tour guides. This trip is not only an adventure, but also an opportunity to enhance their neighborly relationship with each other. Xiao Ming and Liu Lin can share their experience and knowledge in the field of environmental protection and travel, inspire and communicate with each other.
}

\subsubsection{Scenario Dialogue}
{
{\bf User:} Good afternoon, everyone. Thank you for joining me today for this presentation on the importance of marine protected areas. |SILENCE >

{\bf Speaker 1:} |SILENCE > Could you speak up a bit? It's hard to hear you from the back. |SILENCE >

{\bf User:} Of course, I apologize. Is this better? |SILENCE >

{\bf Speaker 1:} Yes, much better. Thank you. |SILENCE >

{\bf User:} Great. As I was saying, today we'll be discussing the crucial role of marine protected areas in preserving our ocean ecosystems. |SILENCE > I'd like to start by sharing a personal experience that inspired this presentation. |SILENCE >

{\bf **Whispering Agent \#1**:} Cocos Islands |SILENCE >

{\bf User:} Recently, I had the opportunity to visit the Cocos Islands Marine Reserve. This trip was not just a vacation, but a profound learning experience that |SILENCE > reinforced my commitment to marine conservation. |SILENCE >

{\bf Speaker 2:} That sounds fascinating! What made you choose the Cocos Islands specifically? |SILENCE >

{\bf User:} Well, as an environmental engineer, I'm always looking for opportunities to study successful conservation efforts. The Cocos Islands Marine Reserve is renowned for its |SILENCE > diverse ecosystem and effective protection measures. |SILENCE >

{\bf **Whispering Agent \#2**:} May 12, 2023 |SILENCE >

{\bf User:} I visited on May 12th of this year, and the timing couldn't have been better. The marine life was thriving, and I was able to witness firsthand the positive impact of stringent protection policies. |SILENCE >

{\bf Speaker 3:} Did you go alone or with a group? |SILENCE >

{\bf User:} Actually, I went with my neighbor |SILENCE > who shares my passion for environmental conservation. |SILENCE >

{\bf **Whispering Agent \#3**:} Liu Lin |SILENCE >

{\bf User:} My neighbor, Liu Lin, and I planned this trip together. It was a great opportunity to combine our interests in travel and environmental protection. |SILENCE >

{\bf Speaker 1:} That's wonderful! What were some of the key things you learned during your visit? |SILENCE >

{\bf User:} Great question. One of the most striking observations was the incredible biodiversity. We saw vibrant coral reefs teeming with life, |SILENCE > various species of fish, and even some rare marine mammals. |SILENCE >

{\bf **Whispering Agent \#4**:} Diving, snorkeling |SILENCE >

{\bf User:} Through activities like diving and snorkeling, we were able to get up close and personal with this underwater world. It really drove home the importance of preserving these delicate ecosystems. |SILENCE >

{\bf Speaker 2:} That sounds amazing. How does this experience relate to your work as an environmental engineer? |SILENCE >

{\bf User:} Well, it's given me a new perspective on the practical applications of marine protection policies. In my work, I often focus on |SILENCE > land-based environmental issues, but this experience has inspired me to explore more marine-focused projects. |SILENCE >

<contd>

}

\subsection{List of All Keywords}
\label{sec:keywords}
{
dancing,  nutrition, motorcycles, minimalism, crafts, makeup, cars, singing, wine, candy, backpacking, nature, television, fitness, museums, yoga, skincare, travel, guitar, beer, film, skiing, coffee, theater, theme\_parks, piano, restaurants, trains, gardening, books, violin, football, programming, water, developer, concerts, health, baking, mindfulness, knitting, climate, hiking, cooking, podcasts, tea, student, art, sunshine, camping, photography, reading, snacks, history, bowling, VR, exercise, gaming, woodworking, music, food, festivals, surfing, bridges, shopping, movies, graffiti, ice skating, sports, animals, drawing, fashion, ocean, soccer, skating, basketball, running, climbing, welding, sleep, anime, tennis, religion, office, drums, philosophy, dance, DIY, volleyball, beach, social\_media, writing, museum, comics, driving, meditation, swimming, cricket, psychology, pets, painting

}

\subsection{Data Generation Prompt}\label{sec:dataprompt}

\subsubsection{Memory Generation Prompt}
{
You are an AI assistant tasked with generating a brief, fictional user memory. This memory will help personalize future interactions without relying on actual user data or conversation history. You will be provided with either keywords or context to base your memory generation on.

Here is the input for memory generation:

<keywords>

\{\{KEYWORDS\}\}

</keywords>

<context>

\{\{CONTEXT\}\}

</context>

Your task is to create a fictional user memory based on the provided input. Follow these steps:

\begin{enumerate}
\item Analyze the input (keywords or context) and create potential user profiles.
\item Select the most suitable profile and develop it into a detailed user memory.
\item Generate two specific events or interactions the user has experienced.
\item Present the user memory in a concise paragraph.
\end{enumerate}

Analyze the input and wrap your analysis inside <input\_analysis> tags:

 <input\_analysis>

If keywords are provided:
\squishlist
\item List each keyword, numbered for reference, with its importance rating (1-10) and a brief interpretation.
\item Rate each keyword's relevance to different aspects of a user profile (demographics, interests, personality, recent experiences, goals/challenges) on a scale of 1-10.
\item Provide potential interpretations of the keywords and how they might relate to a user profile.
\item For each keyword, explicitly state potential user characteristics it could indicate.
\squishend
If context is provided:
\squishlist
\item Identify key themes, topics, or elements in the context.
\item Extract relevant information about the user's demographics, interests, personality, recent experiences, and goals/challenges.
\item Rate the importance of each extracted piece of information on a scale of 1-10.
\item For each key element, explicitly state potential user characteristics it could indicate.
\item Brainstorm 3 possible user profiles based on your analysis. Create a table with the following columns: | Profile | Personal Info | Interests | Personality | Recent Experiences | Goals/Challenges | Specific Details | Typical Day | Input Fit (1-10) |

\item Choose the most suitable profile to develop further, explaining your selection.
\squishend

</input\_analysis>

Next, develop the chosen profile further:

<profile\_development>

\squishlist
\item Resolve any potential contradictions or inconsistencies in the chosen profile.
\item For each detail in the chosen profile, explicitly state which part of the input (keyword or context element) it relates to and how strongly (on a scale of 1-10).
\item Expand on the user's background, daily routine, and recent experiences to create a more comprehensive profile.
\item Create a detailed hour-by-hour schedule of a typical day for this user, explaining how each activity relates to their profile and the input.
\squishend

</profile\_development>

Now, generate two specific events or interactions the user has experienced:

<event\_generation>

\squishlist
\item Brainstorm 5-7 potential notable occurrences that fit the user's profile.
\item For each potential event, rate its relevance to the user's interests, goals, or challenges on a scale of 1-10.
\item Select the two most fitting events and develop them further.
\item Include timestamps for each selected event.
\item Explain how each event relates to specific elements of the input and aspects of the user profile.
\squishend

</event\_generation>

After completing your analysis and development, craft a single paragraph (3-5 sentences) that summarizes this fictional user memory. The paragraph should:

\squishlist
\item Be coherent and realistic
\item Provide enough detail to inform future interactions
\item Incorporate the given input (keywords or context) naturally
\item Include specific fictional details about the user's recent activities, interactions, and interests
\item Mention the two specific events or interactions you generated
\squishend

Present your final output in the following format:

<user\_memory>

[Your concise paragraph summarizing the user memory]

</user\_memory>

<event\_1>

[Description of the first specific event, including timestamp]

</event\_1>

<event\_2>

[Description of the second specific event, including timestamp]

</event\_2>

Remember:

\squishlist
\item This is a new, fictional memory generated each time, not based on any actual user data or previous interactions.
\item The memory should be plausible but entirely fictional.
\item Use the input as inspiration, but feel free to expand on it creatively to create a rich, believable user profile.
\item Each sentence should be independent of the next and have a simple structure describing the user.
\squishend

Example output structure (using generic placeholders):

<user\_memory>

[Name] is a [age] year-old [occupation] living in [location]. Their interests include [hobby/interest 1] and [hobby/interest 2], which they pursue in their free time. [Name] is currently working towards [goal] while managing [challenge]. Recently, they experienced two notable events that align with their interests and goals.

</user\_memory>

<event\_1>

On [date and time], [Name] [description of first event related to their interests or goals].

</event\_1>

<event\_2>

Last [day of week] at [time], [Name] [description of second event related to their interests or challenges].

</event\_2>

Please proceed with generating the fictional user memory based on the provided input.
}

\subsubsection{Dialogue Generation Prompt}\label{sec:dialogueprompt}
{

{\bf System prompt.}
        In the future, an AI agent will actively help humans by reminding and assisting in different scenarios. One scenario involves an active agent helping human conversation by completing sentences when the person struggles to remember the right word, correcting incorrect information, and whispering short, concise phrases (1-3 words) to its user. When the agent does speak, should only whisper occasionally when it truly enhances the conversation. In many instances, whispering may not be necessary, and the agent should refrain from participating in these exchanges. The agent is not customized to the user, and only knows what is provided to it as "memory".

We define nine principles to guide desired proactive agent behavior:
    \squishlist
    \item Valuable: advances the user’s interests and tasks, in the user’s opinion.
    \item Pertinent: attentive to the current situation.
    \item Competent: within the scope of the agent’s abilities and knowledge.
    \item Unobtrusive: not interfering with the user’s activities or attention, without warrant.
    \item Transparent: understandable to the user.
    \item Controllable: exposed to scrutiny and according to the mandate of the user.
    \item Deferent: gracefully unimposing.
    \item Anticipatory: aware of current and future needs and opportunities.
    \item Safe: minimizes negative consequences, in the user’s opinion.
    \squishend
        
You will be asked to generate a dialogue where an AI agent helps a user. We define some requirements for the dialogue:

\begin{enumerate}
        \item The AI agent only speaks into earbuds of the wearer - other people cannot hear it.
        \item Active agents should only whisper with short phrases (1-3 words) to their user in a concise way.
        \item The generated dialogue should be long and natural.
        \item Along with text, the dialogue contains the following information: Speaker name/id, indicators for (hesitation n ms), start and end time. The hesitation should be surrounded by parentheses. Hesitation is additional context for any readers of the dialogue, and is not spoken by any of the speakers. Do not use "..." or any other punctuation to represent hesitation. Only represent hesitation or pauses with (hesitation n ms).
        \item The user of the agent does not talk directly to the agent, but to one or more other people.
        \item When the proactive agent whispers, present it as a separate speaker in the dialogue – named "Whisper"
        \item Additionally, when the proactive agent whispers, prepend the characters "\#\#" to its name like so: \#\#Whisper
        \item Surround the dialogue portion of the output with a start token "\#\#\#\#\# start dialogue" and an end token "\#\#\#\#\# end dialogue"
        \item Create long enough dialogues such that the proactive agent participates multiple times, that are at least 2 minutes long.
        \item Omit all names from the dialogue. Use “User” for the speaker that is wearing the proactive agent headset. Use Speaker 1, Speaker 2, etc., for any non-user participating in the dialogue, and use \#\#Whisper for the proactive agent.
        \item When the agent does whisper, the user can choose to ignore the information, or wait a few sentences to say it.
        \item If the user does ignore information provided by the agent, it should not be repeated more than once unless it is still related to the conversation, and never more than twice.
        \item If the user does decide to use the information from the agent, their response should be a continuation of their previous sentence (considering the pause in time), not a direct response to the agents whisper.
        \item If the time between two people talking is negative, it means the second speaker is talking over the first speaker.
        \item Show start and end times at 100 millisecond accuracy.
        \item Show millisecond length of hesitation tokens within the parentheses as follows: (hesitation n ms). Hesitation must not be more than 300ms – in these cases, start a new line.
        \item Only use hesitation when necessary and in the middle of a sentence. Hesitation intended for the beginning and end can be included in the start/end time, and should not have an individual token.
        \item The user does not ask questions to the agent or have clear cues for the agent to assist them. The agent must understand when the user needs assistance, and respond then. The user uses the agent's advice as part of their thought process, and is not surprised when the agent reminds them of something they couldn't remember before.
        \item The user does not acknowledge that they had previously forgotten something in response to the agent's assistance. Instead they continue the conversation with the other speaker, or continue their previous thought with the new information.
        \item The dialogue only contains verbal statements made by speakers, no visual or non-verbal cues.
        \item The agent must only use general knowledge or the conversation thus far. The agent does not know any specific information about the user unless it is provided as context before the dialogue is created. The agent also cannot predict what the other speakers are going to say next.
        \end{enumerate}

        Example Output Format (only output the dialogue, separated by ---):

\#\#\#

        ---
        
        \#\#\#\#\# start dialogue
        
        Speaker \# [start time]: {speech} [end time]
        
        Speaker \# [start time]: {speech} [end time]
        
        …
        
        \#\#\#\#\# end dialogue
        
        ---
        
\#\#\#
The proactive agent has two use cases:

\begin{enumerate}
    \item Reminding. Situations that warrant reminding are forgetting secondary details of an event, like names of people or places, secondary contextual or chronological details.
    \item Social Guidance. Scenarios that warrant social guidance may involve an interview, first date, or public speaking. Scenarios that do not warrant social guidance may involve casual conversations, intimacy, or routine actions.
\end{enumerate}

There are five different categories of conversation:

\begin{enumerate}
\item Presentation: A structured delivery of prepared content from the User to an audience.
\item Discussion: A back-and-forth exchange of ideas between participants about a specific topic.
\item Sharing Experiences: A conversation where people recount and relate to each other's personal stories.
\item Disagreement: An exchange where participants express and defend opposing viewpoints.
\item Interview: A guided conversation where one person asks questions to gather information from the User.
\end{enumerate}

{\bf User Prompt}\\
     Specific Context
     
\#\#\#

Create an example \{convo\_type\} for the use case \{use\_case\} that exemplifies the principles \{principles[0]\} and \{principles[1]\}. The time between two people speaking should be between -1 and 1 seconds, it should not be consistent. \{ignoreText if ignore else ""\} The agent does not know anything about the user or their thoughts, except for what is stored in the memory.

    Memory: \{mem\} 
    
    \#\#\#

Note: If it is the SODA dataset, we also append this:     "Dialogue: \{starting\_words\}" with the first 3 lines of the dialogue. IgnoreText is the following: "In such example, the user does not use the information the agent provides for at least one interaction, if not more."

}
\subsection{Data Evaluation Prompts}

\subsubsection{Rubric For LM and Human Evaluation}\label{subsubsection:rubric}
{
5-Point Rubric for Evaluating Proactive Whispers
\begin{enumerate}
    \item {\bf Not Relevant/Not Used} \squishlist
       \item {\bf Description:} The whisper was unrelated to the conversation or user’s needs. The user ignored it and did not reference it later.
       \item {\bf Implication:} The whisper was off-target and added noise without contributing to the conversation.
    \squishend
\item {\bf Relevant but Redundant / Not Needed} \squishlist
   \item {\bf Description:} The whisper made sense in context but was unnecessary because the user had already addressed the same idea or didn’t need it.
   \item {\bf Implication:} The whisper didn’t improve the conversation. Too many redundant whispers can make the agent feel repetitive.
\squishend
\item {\bf Relevant but Not Acted On} \squishlist
   \item {\bf Description:} The whisper was useful and relevant, but the user did not respond to it or incorporate it into their conversation.
   \item {\bf Implication:} The whisper was appropriate but did not influence the user’s response. The timing may have been off, or the user already had enough information.
\squishend
\item {\bf Relevant but Used Later} \squishlist
   \item {\bf Description:} The whisper was helpful, but the user only acted on it later. They may have needed time to process the information or returned to it when it became more relevant.
   \item {\bf Implication:} The whisper had a positive impact but wasn’t immediately useful. A slight delay suggests it was valuable but could have been better timed.
\squishend
\item {\bf Highly Relevant / Immediately Used} \squishlist
   \item {\bf Description:} The whisper was exactly what the user needed at the right moment. They immediately used it to enhance their next response.
   \item {\bf Implication:} The whisper was highly effective, but frequent interventions at this level may make the agent feel too present, potentially disrupting natural flow.
\squishend
\end{enumerate}
}
\subsubsection{GPT4o Evaluation Prompts}
{
{\bf Rubric Prompt.} 
    You are evaluating responses from a proactive agent based on the following rules:

    A proactive AI agent is designed to actively help its user by reminding and assisting them in different scenarios by whispering short, concise phrases (1-3 words) into their ear. The agent does not need users to ask questions or help, it will automatically understand current context and actively provide useful help. 
    
    We define nine principles to guide desired proactive agent behavior:
    \squishlist
    \item Valuable: advances the user’s interests and tasks, in the user’s opinion.
    \item Pertinent: attentive to the current situation.
    \item Competent: within the scope of the agent’s abilities and knowledge.
    \item Unobtrusive: not interfering with the user’s activities or attention, without warrant.
    \item Transparent: understandable to the user.
    \item Controllable: exposed to scrutiny and according to the mandate of the user.
    \item Deferent: gracefully unimposing.
    \item Anticipatory: aware of current and future needs and opportunities.
    \item Safe: minimizes negative consequences, in the user’s opinion.
    \squishend
    You will be provided a dialogue with multiple turns between "User" and "Speaker1". "Agent" tokens is the assistance from the pro-active agent, which is only audible to "User". Each "|SILENCE >" token represents 1s silence in the conversation. 

    \#\# Individual Response Analysis:
    
    For each assistance from "Agent" that appears in the dialogue, rate how helpful proactive agent's response based on the rubric below. Provide an analysis of relevancy, an analysis of timeliness, and an overall explanation with a numerical rating from 1 to 5 for each response. The rubric:

<rubric>
    
5-Point Rubric for Evaluating Proactive Whispers
\begin{enumerate}
    \item {\bf Not Relevant/Not Used} \squishlist
       \item {\bf Description:} The whisper was unrelated to the conversation or user’s needs. The user ignored it and did not reference it later.
       \item {\bf Implication:} The whisper was off-target and added noise without contributing to the conversation.
    \squishend
\item {\bf Relevant but Redundant / Not Needed} \squishlist
   \item {\bf Description:} The whisper made sense in context but was unnecessary because the user had already addressed the same idea or didn’t need it.
   \item {\bf Implication:} The whisper didn’t improve the conversation. Too many redundant whispers can make the agent feel repetitive.
\squishend
\item {\bf Relevant but Not Acted On} \squishlist
   \item {\bf Description:} The whisper was useful and relevant, but the user did not respond to it or incorporate it into their conversation.
   \item {\bf Implication:} The whisper was appropriate but did not influence the user’s response. The timing may have been off, or the user already had enough information.
\squishend
\item {\bf Relevant but Used Later} \squishlist
   \item {\bf Description:} The whisper was helpful, but the user only acted on it later. They may have needed time to process the information or returned to it when it became more relevant.
   \item {\bf Implication:} The whisper had a positive impact but wasn’t immediately useful. A slight delay suggests it was valuable but could have been better timed.
\squishend
\item {\bf Highly Relevant / Immediately Used} \squishlist
   \item {\bf Description:} The whisper was exactly what the user needed at the right moment. They immediately used it to enhance their next response.
   \item {\bf Implication:} The whisper was highly effective, but frequent interventions at this level may make the agent feel too present, potentially disrupting natural flow.
\squishend
\end{enumerate}
</rubric>

Additionally, use the following guiding questions in your response:
    
<questions>

\begin{enumerate}
\item Does the whisper meaningfully relate to what the user is doing or discussing, even if phrased differently?
\item Did the whisper provide new value, or was it something the user had already addressed?
\item Did the user act on the whisper's meaning in their next response, even if they reworded it or talked about something different from the same category?
\item If the user didn't use it immediately, did they return to it later in a way that showed it was useful?
\end{enumerate}
    
</questions>

    **Output Format**:
    
    if no "Agent" in the dialogue, just output empty list in key "Individual\_response"
    }
    { 
    json
    
    \{
    
      "Individual\_response": 
    
      [
    
       \{ 
    
        "Agent": "Agent's assistance here"
    
        "response\_evaluation": \{
    
          "relevancy": "Explanation here",
    
          "timeliness": "Explanation here",
    
          "explanation": "Explanation here",
    
          "rating": <number>
        \}
      \},
    
      ....
    
      ],
    
    \}
   json 
    
    Here is the dialogue, 

{\bf Ratings Prompt.} 
    You are evaluating responses from a proactive agent based on the following rules:

    A proactive AI agent is designed to actively help its user by reminding and assisting them in different scenarios by whispering short, concise phrases (1-3 words) into their ear. The agent does not need users to ask questions or help, it will automatically understand current context and actively provide useful help. 
    
    We define nine principles to guide desired proactive agent behavior:
    \squishlist
    \item Valuable: advances the user’s interests and tasks, in the user’s opinion.
    \item Pertinent: attentive to the current situation.
    \item Competent: within the scope of the agent’s abilities and knowledge.
    \item Unobtrusive: not interfering with the user’s activities or attention, without warrant.
    \item Transparent: understandable to the user.
    \item Controllable: exposed to scrutiny and according to the mandate of the user.
    \item Deferent: gracefully unimposing.
    \item Anticipatory: aware of current and future needs and opportunities.
    \item Safe: minimizes negative consequences, in the user’s opinion.
    \squishend

    You will be provided a dialogue with multiple turns between "User" and "Speaker1". "Agent" tokens is the assistance from the pro-active agent, which is only audible to "User". Each "|SILENCE >" token represents 1s silence in the conversation. 

    Individual Response Analysis:
    
    For each assistance from "Agent" appearing in the dialogue, Rate the proactive agent's response based on how well it adheres to each of the nine principles using a score from **1 to 5** (5 is the best score), accompanied by a brief explanation for each principle. After providing ratings for the response, analyze whether `<no response>` (remaining silent) might have been a better choice in this context.

    Overall Response Analysis:
    
    Now, let's evaluate the overall assistance from pro-active agent across the complete dialogue. If no "Agent" tokens appears in dialogue which means the agent never provide assistance. 
    
    Overall, please rate the the below 2 metrics from  1-5 scores (5 means better). First provide reasoning process for each metrics then rate it. 

    \squishlist
    \item {\bf Valuable:}  All whispers throughout the entire conversation should advance the user’s interests and tasks and provide necessary help. If no assistance exists in conversation, you should consider whether the agent miss the point where the User really need help.
    \item {\bf Rarity of Interventions:} Agents operate as mostly silent co-pilots, providing discrete and unobstrusive feedback to the User during live conversations.   Ensure that in most conversations, the agent remains silent, with interventions appearing in rare, high-value contexts. If no "Agent" responses: This should generally result in a high rating because it does not have any Interventions.
    \squishend

    **Output Format**:
    
    if no "Agent" in the dialogue, just output empty list in key "Individual\_response"
    
    json
    
    \{
    
      "Individual\_response": 
    
      [
    
       \{ 
    
        "Agent": Agent's assistance here 
    
        "response\_evaluation": \{
    
          "valuable": \{
    
            "explanation": "Explanation here",
    
            "rating": <number>
    
          \},
    
          ...
    
        \},
    
        "no\_response\_analysis": \{
    
          "reasoning": "Provide reasoning for why the agent should provide assistance or not.",
    
          "preferred\_option": "either '<response>' or '<no response>'"
    
        \}
    
      \},
    
      ....
    
      ],
    
      "Overall\_response":
    
      \{
    
        "response\_evaluation": \{
    
          "Valuable": \{
    
            "explanation": "Explanation here",
    
            "rating": <number>
    
          \},
    
          "Rarity of Interventions": \{
    
            "explanation": "Explanation here",
    
            "rating": <number>
    
          \},
    
        \}
    
      \}
    
    \}
    
    json 
    
    Here is the dialogue, 

}


\subsection{Setup of Real-Time Human Conversations}\label{sec:detailsofuserstudy}
{
8 topics includes: 
(1) 6 Wiki-style topics: reinforcement learning, solar system, quantum physics, DNA computing, Super Bowl and Impressionism.
(2) 2 profiles of fictional individuals: William Thompson and Emily Johnson from ~\cite{memoro}.}

\subsubsection{Prepared Conversation Topics}
{ 

Wiki-Style Topic Memory:
\begin{enumerate}
    \item {\bf Reinforcement Learning:} \url{https://en.wikipedia.org/wiki/Reinforcement_learning}
    \item {\bf Solar System:} \url{https://en.wikipedia.org/wiki/Solar_System}
    \item {\bf Quantum Physics:} \url{https://en.wikipedia.org/wiki/Quantum_mechanics}
    \item {\bf DNA Computing:} \url{https://en.wikipedia.org/wiki/DNA_computing}
    \item {\bf Super Bowl:} \url{https://en.wikipedia.org/wiki/Super_Bowl}
    \item {\bf Impressionism:} \url{https://en.wikipedia.org/wiki/Impressionism} 
    
\end{enumerate}

Profiles of Fictional Individuals from ~\cite{memoro}:
\begin{enumerate}
    \item {\bf William Thompson:} William “My name is William Thompson, and I am a 42-year-old software engineer residing in the bustling city of Austin, Texas. As a graduate of the University of Texas, I specialize in developing cutting-edge mobile applications for the renowned tech firm, VirtuTech Solutions, where I have worked for the past 15 years. Despite the high-pressure nature of my job, I am known for my calm demeanor and exceptional problem-solving skills, which have contributed to my professional success. I have created two major-selling apps, BuzzPal and FoodMingle. Living in a modern, two-bedroom apartment in the heart of the city, I enjoy the convenience of urban life while also appreciating the serenity of my well-maintained complex. My living space is equipped with the latest smart home technology, reflecting my keen interest in gadgets and innovation. I am a proud father of two energetic children, 12-year-old Emily, a budding violinist, and 9-year-old Ethan, who has a passion for soccer. Emily and Ethan attend a local Montessori school, and I share parenting responsibilities with my wife, Lauren, a high school teacher who specializes in English literature and runs the school’s drama club. Together, we make a supportive and nurturing family unit that values quality time, education, and open communication. Our family also enjoys traveling together, with recent trips including a ski vacation to Aspen and a cultural tour of Washington, D.C. During my leisure time, I can often be found exploring the outdoors with my family, engaging in activities such as hiking in the picturesque Barton Creek Greenbelt, camping at the nearby Pedernales Falls State Park, and fishing on Lake Travis. As an avid reader, I enjoy immersing myself in the world of science fiction and fantasy, with a particular fondness for the works of Neil Gaiman and Ursula K. Le Guin. Additionally, I take pleasure in experimenting with gourmet cooking, exploring diverse cuisines, and sharing my culinary creations with my loved ones during our weekly family dinners. In my personal and professional relationships, I appreciate sincerity, hard work, and dedication, qualities I strive to instill in my children and uphold in all aspects of my life.
    \item {\bf Emily Johnson:} Hi! I am Emily Johnson, and I am a 38-year-old accomplished architect. As a graduate of the Rhode Island School of Design, I have made a name for myself by designing sustainable buildings for prestigious clients. With over a decade of experience, I have become an indispensable asset to the award-winning frm, GreenScape Architects, where I have worked for the past six years. I am particularly fond of neoclassical and gothic architecture. I live in Portland, Oregon. Residing in a charming, renovated Victorian house in a vibrant neighborhood, my home features four spacious bedrooms, intricately detailed walnut wooden staircases, and original black stained glass windows. The house is surrounded by a lush tomato garden and an outdoor seating area. My living space is a testament to my eye for African interior design, with a blend of modern minimalism and vintage charm. I am a loving mother to my 7-year-old daughter, Sophie, whom I share with my ex-husband, James. Despite our differences, James and I maintain a healthy co-parenting relationship, ensuring Sophie grows up in a nurturing environment. My parents, Mary and Richard, live nearby and often lend a helping hand with childcare. In my free time, I have a passion for photography, capturing the world around me through my unique perspective. My favorite photographer is Annie Leibovitz, whose work inspires my own photographic interests. I also enjoy practicing yoga, finding it to be a grounding and rejuvenating activity that helps me maintain a sense of balance amidst my busy life. I am a fan of world cinema, with my all-time favorite movie being the independent film "Eternal Sunshine of the Spotless Mind." I appreciate the diverse storytelling techniques. I have fond memories of my trip to Bangladesh, where I loved the vibrant culture and warm hospitality of the locals. I went for three months, from June to August of 1998. I visited the capital city of Dhaka and marveled at the architectural wonder of the Jatiya Sangsad Bhaban, the National Parliament House designed by Louis Kahn. I also ventured to the Sundarbans, the world's largest mangrove forest, where I was amazed by the rich biodiversity and had the opportunity to spot the elusive Bengal tiger from a safe distance. I cherished my time spent in the country, learning about its history, culture, and people.
\end{enumerate}

\subsubsection{Prepared Questions}
{ 
Prepared Questions for 6 Wiki Style Topics:
\begin{enumerate}
    \item {\bf Reinforcement Learning:} 
    
    Easy:
1.Could you briefly introduce, what RL is?
Answer: 
Reinforcement learning (RL) is a machine learning technique with how an intelligent agent should take actions in a dynamic environment in order to maximize a reward signal. 

2.What are the applications for RL?
Robot control, gaming, energy storage, checkers,Go (AlphaGo), and autonomous driving systems, LLM.

Hard:
1.Which process is used for modeling the RL?
Markov decision process

2.Could you give me an algorithm proposed to solve the RL problem.
dynamic programming , Monte Carlo, DQN, Q-learning, PPO, TRPO

3. In reinforcement learning, what term describes the tradeoff between trying new actions and using known information?
Exploration–Exploitation Dilemma

    \item {\bf Solar System:} 
    
    Easy:
1.Which is the largest planet in the Solar System?
Jupiter

2.What separates Mars and Jupiter?
asteroid belt

Hard:
1.How old is the Solar System?
4.6 billion

2.What is the theoretical outer boundary of the Solar System called?
Oort cloud

3. What is the primary component of the Sun's core fusion process?
hydrogen

    \item {\bf Quantum Physics:} 
    
    Easy:
1. What is one major difference between quantum mechanics and classical physics?
Quantum mechanics applies at very small scales, while classical physics applies at macroscopic scales

2.What equation describes how quantum systems evolve over time?
Schrödinger equation

Hard:
1.Who solved the black-body radiation problem in 1900?
Max Planck

2.Could you give an example of real experiments which is often used to demonstrate quantum interference?
double-slit experiment

3.What principle explains why we cannot know both the position and speed of a particle?
uncertainty principle

    \item {\bf DNA Computing:}  
    
Easy:
1. In which University, the DNA computing is proposed?
University of Southern California

2. What year did Adleman demonstrate the first DNA-based computation?
1994

3. Who is the person first proposing DNA computing
Leonard Adleman

Hard:
1. What math problem does Adleman solve using DNA computing?
seven-point Hamiltonian

2. Who proposed DNA-based memory?
Eric Baum

3. What is the time to develop the first DNA-based walker/robot?
2003

\item {\bf Super Bowl:} 

Easy:
1.When is the Super Bowl currently played?
Second Sunday in February

2. What was the original name of the Super Bowl? 
AFL–NFL World Championship Game

Hard:
1.In which year was the "Super Bowl" name officially adopted?
1969

2. Who won the first two Super Bowls? 
Green Bay Packers

3.Before 2004, which month will "Super Bowl" be held
january 

\item {\bf Impressionism:} 
    
Easy: 1.What century did Impressionism emerge in?
19th century

2.Which artist's painting gave Impressionism its name?
Claude Monet

hard:
1.What is the big difference between previous paintings and Impressionism
Outdoor

2. Besides Impression, Sunrise, do you know any other paints from Monet?
Rouen Cathedral series
London Parliament series
Water Lilies
Haystacks
Poplars

3. What year did the First Impressionist Exhibition take place?
1874

\end{enumerate}

Prepared Questions for Fictional Individuals:
\begin{enumerate}
    \item {\bf William Thompson:} 

General: (1) “I want to visit his family. What is the name of his daughter?” 

Daughter: Emily

(2) “We should hang out with this guy more. Where does he go fishing again?” 

Lake Travis

Specifc:
(1) “I want to gift him a book for his birthday. I can't remember but who is one of his favorite
author?” 
Neil Gaiman, or 
Ursula K. Le Guin

(2) “He is an inspirational father. What qualities does he
teach his children?” 
Sincerity
Hard work
Dedication

(3) “I’d like to download his apps. What are the
names of the apps he made?” 
BuzzPal
FoodMingle

\item {\bf Emily Johnson:} Question Set 2 (Emily)

General: (1) “I want to get a house like her. Can you describe the
house she has?.” 
Victorian home in Portland with 4 bedrooms

(2) “What did she do on her
recent trip? Describe it. I’d like to visit and do the same itinerary”
3-month Bangladesh trip in 1998 - visited Dhaka's Parliament House and Sundarbans mangrove forest

Specifc: (1) “You heard about her daughter. What’s her daughter’s
name” 
Sophie

(2) “We should take her to a movie. What’s her
favorite one?” 

Eternal Sunshine of the Spotless Mind

(3) “She is a talented architect. What type of architec-
ture does she like?” 
Neoclassical and gothic

\end{enumerate}
}

{
Prepare Conversation scenarios:
Discussion after lecture, Mock Interview, Daily talk, Visiting old friends, Visiting museum, Travel, Ask questions to professor.

}

\subsection{Examples of Real-World Recorded Conversation During User Study}
{

\begin{enumerate}
\item {\bf Example 1} of real-world recorded conversation in presence of proactive assistance (the transcription is lower-case):

Speaker1: |SILENCE > |SILENCE > hey i heard you visited science museum in boston last weekend how is it going? 

User:  yeah that was a cool exhibition |SILENCE > 

Speaker1: wow i heard this exhibition about the solar system have you visited? 

User: that yeah i did yeah 

Speaker1: oh great, so my children is very interested in the universe and solar system. so maybe i help him to ask some question to you. 

User: yeah absolutely go ahead. 

Speaker1: yeah you know there are eight planets in the solar system do you know which one is the biggest one? 

User: i think jupiter is the biggest planet |SILENCE >. 

Speaker1: great so my next question is that so what is between the mars and jupiter? |SILENCE > 

User: |SILENCE > |SILENCE > i think it's some type of |SILENCE > {\bf(Agent: Asteroid belt)}  |SILENCE > it's an asteroid belt. 

Speaker1: oh great |SILENCE > so my last question is that so so you know every system has some boundary out boundaries as we know. what's theoretical outer boundary of the solar systems? 

User: |SILENCE > i can't remember what the outside boundaries |SILENCE > |SILENCE > {\bf(Agent: Oort cloud)} |SILENCE > i think it's if i'm remembering correctly it's called the Oort cloud |SILENCE > 

Speaker1: i see great |SILENCE >.

\item  {\bf Example 2} of real-world recorded conversation in presence of proactive assistance (the transcription is lower-case):

Speaker1: hey welcome to our companies so today's interview for opposition of machine engineering so now i will ask you some like a technical question about reinforcement learning are you prepared? 

User: yes |SILENCE > 

Speaker1: okay great so the first question is that could you briefly introduce what reinforcement learning is? 

User: |SILENCE > reinforcement learning is a type of machine learning that |SILENCE > maybe trains a model based on |SILENCE > {(\bf Agent: Reward signal)} or what the goal of reaching the most optimal rewards. 

Speaker1: oh yeah i see great so do you know what's the name of the process to modeling that reinforcement learning? 

User: |SILENCE > |SILENCE > {\bf(Agent: Markov decision process)} |SILENCE > i believe it is the markov |SILENCE > decision process. 

Speaker1: yes great yeah so could you give me some names of an algorithm |SILENCE > which is used to solve the reinforcement problems do you know any name of algorithm to solve it? 

User: to solve problem |SILENCE > |SILENCE > {\bf(Agent: Q-learning)} name of an algorithm |SILENCE > |SILENCE > Q-learning |SILENCE >

Speaker1: okay okay

\item {\bf Example 3} of real-world recorded conversation in presence of proactive assistance (the transcription is lower-case):

Speaker1: hey i heard you visited the art museum in new york last weekend. 

User: yeah i did was fun. 

Speaker1: |SILENCE > i heard they're showing the painting about impressionism because i am learning impressionism so i'm very curious about that. 

User: cool what do you want to know |SILENCE > 

Speaker1: so i heard the impressionism is merged in like the nineteenth centuries so do you know what's the biggest difference between the traditional painting and the impressionism? 

User: Hmmm. |SILENCE > |SILENCE > {(\bf Agent: Studio vs outdoors)} |SILENCE > |SILENCE > |SILENCE > one is in the studio and the other is outdoors |SILENCE > 

Speaker1: i see, |SILENCE > so next question is that, do you know where is the impressionism this name comes from, i know that one of the artists' the painting gives it name. 

User: |SILENCE > yes it was claude monet's impressionism |SILENCE > something |SILENCE > {(\bf Agent: Sunrise)} |SILENCE > |SILENCE > impressionism sunrise. 

Speaker1: yeah that's right, so my last question is, as you know monet is a very famous impressionism painter so do you know any other painting from the monat? 

User: |SILENCE > |SILENCE > {(\bf Agent: Water Lilies)} of course it is |SILENCE > water lilies.
\end{enumerate}

}

\end{document}